\documentclass[11pt,a4paper]{article}

\usepackage[utf8]{inputenc}
\usepackage[T1]{fontenc}
\usepackage[english]{babel}
\usepackage{amsmath}
\usepackage{amssymb}
\usepackage{graphicx}
\usepackage{booktabs}
\usepackage{hyperref}
\usepackage{url}
\usepackage{geometry}
\usepackage{float}
\usepackage{xcolor}
\usepackage{caption}
\usepackage{subcaption}
\usepackage{array}
\usepackage{multirow}
\usepackage{setspace}
\usepackage{parskip}
\usepackage{lscape}
\usepackage{pgfplots}
\usepackage{pgfplotstable}
\pgfplotsset{compat=1.18}

\hypersetup{colorlinks=true, linkcolor=blue, citecolor=blue, urlcolor=blue}

\makeatletter
\renewcommand{\@maketitle}{%
  \newpage\null\vskip 2em%
  \begin{center}%
    {\LARGE \@title \par}\vskip 1.5em%
    {\large \@author \par}\vskip 1em%
    {\large \@date}%
  \end{center}\par\vskip 1.5em}
\makeatother

\title{\textbf{Beyond Acoustic Emotion Recognition:\\
Multimodal Pathos Analysis in Political Speech\\
Using LLM-Based and Acoustic Emotion Models}}

\author{\begin{center}
  J\"{u}rgen Dietrich\textsuperscript{\href{https://orcid.org/0000-0002-5494-3499}{0000-0002-5494-3499}}\\[0.3em]
  Democracy Intelligence gGmbH, Germany\\
  \texttt{juergen.dietrich@democracy-intelligence.de}
  \end{center}}

\date{May 2026}

\begin{document}
\maketitle

\begin{abstract}
We investigate whether acoustic emotion recognition models can serve as proxies
for the Pathos dimension in political speech analysis, as operationalised by the
TRUST multi-agent large language model (LLM) pipeline. Using a Bundestag plenary
speech by Felix Banaszak (51 segments, 245\,s) as a case study, we compare three
analysis modalities: (1)~\textit{emotion2vec\_plus\_large}, an acoustic speech
emotion recognition (SER) model whose continuous Arousal and Valence values are
derived via post-hoc Russell Circumplex projection; (2)~\textit{Gemini 2.5
Flash}, an LLM analysing the full speech audio together with its transcript in
an open-ended, context-aware fashion; and (3)~TRUST-Pathos scores from a
three-advocate LLM supervisor ensemble. Spearman rank correlations reveal that
Gemini Valence correlates strongly with TRUST-Pathos ($\rho = +0.664$,
$p < 0.001$), whereas emotion2vec Valence does not ($\rho = +0.097$,
$p = 0.499$). We further demonstrate, via a systematic quality evaluation of
the Berlin Database of Emotional Speech (EMO-DB) using Gemini in an open-ended
annotation paradigm, that standard SER benchmark corpora suffer from acted
speech, cultural bias, and category incompatibility. Our results suggest that
LLM-based multimodal analysis captures semantically defined political emotion
substantially better than acoustic models alone, while acoustic features remain
informative for low-level Arousal estimation. Future work will extend this
approach to video-based analysis incorporating facial expression and gaze.

\end{abstract}

\clearpage

\section{Introduction}

The analysis of emotional expression in political discourse has received increasing attention from communication scholars, political scientists, and computational linguists. Within the TRUST framework~\cite{dietrich2026trust,dietrich2026roles},
political statements are evaluated along three rhetorical dimensions inspired by
Aristotelian rhetoric: Logos (logical argumentation), Ethos (credibility), and
Pathos (emotional appeal). While Logos and Ethos can be assessed through
structured fact-checking and credibility scoring, Pathos poses a particular
challenge: it requires the recognition of affective states embedded not only in
lexical content but also in prosody, rhythm, and rhetorical strategy.

Automatic SER has advanced substantially in recent years, with self-supervised
models such as \textit{wav2vec\,2.0}~\cite{baevski2020wav2vec} and dedicated
emotion encoders like \textit{emotion2vec}~\cite{ma2023emotion2vec} achieving
competitive performance on standard benchmarks. However, these benchmarks are
dominated by acted corpora---most prominently the Interactive Emotional Dyadic
Motion Capture database (IEMOCAP)~\cite{busso2008iemocap} and
EMO-DB~\cite{burkhardt2005database}---whose ecological validity for naturalistic
political speech remains unexplored.

In parallel, LLMs with multimodal capabilities have begun to demonstrate
remarkable competence in understanding emotional nuance, irony, and rhetorical
intent from audio and text. This raises the question of whether LLM-based
analysis can serve as a more valid proxy for the politically relevant Pathos
dimension than acoustic SER.

This paper makes three main contributions:
\begin{enumerate}
  \item We compare emotion2vec Arousal/Valence, Gemini multimodal
        Arousal/Valence, and TRUST-Pathos scores on 51 segments of a
        naturalistic Bundestag (German Federal Parliament) speech, demonstrating that LLM-based emotion
        analysis correlates substantially better with LLM-based Pathos scoring
        than acoustic SER.
  \item We provide a systematic quality evaluation of EMO-DB using Gemini as an
        open-ended, non-forced-choice annotator, revealing structural limitations
        of this widely used corpus, including undocumented discrepancies in
        gender coding and sentence transcription (see Appendix~A).
  \item We introduce the concept of \textit{post-hoc Russell Circumplex
        projection} as an operationalisation of continuous Arousal/Valence from
        discrete SER class probabilities, and discuss its assumptions and
        limitations.
\end{enumerate}

The remainder of this paper is structured as follows.
Section~\ref{sec:related} reviews related work.
Section~\ref{sec:methods} describes our methods and data.
Section~\ref{sec:results} presents all results, comprising the EMO-DB quality
evaluation (Section~\ref{sec:emodb}) and the Banaszak speech analysis
(Section~\ref{sec:banaszak}).
Section~\ref{sec:discussion} discusses implications and limitations.
Section~\ref{sec:conclusion} concludes.

\section{Related Work}
\label{sec:related}

This section situates our work within three research streams: automatic SER,
multimodal LLM-based affect analysis, and computational political communication.

\subsection{Speech Emotion Recognition}

The field of SER has been shaped by a small number of acted corpora. EMO-DB~\cite{burkhardt2005database},
recorded in 1999 at the Technische Universit\"{a}t Berlin, comprises 535
utterances from ten professional actors (five female, five male) covering seven
emotion categories across ten fixed German sentences. Despite its age, EMO-DB
remains a standard benchmark. The EmoBox framework~\cite{ma2024emobox}, which
evaluates ten pre-trained speech models across 32 datasets in 14 languages,
reports that WavLM Large achieves 92.67\,\% weighted accuracy (WA) on EMO-DB,
while wav2vec\,2.0 base reaches 83.14\,\% WA.

The emotion2vec model family~\cite{ma2023emotion2vec} employs self-supervised
pre-training on 262 hours of open-source emotional speech data (internally
termed \textit{Emo-262}), followed by fine-tuning. The composition of Emo-262
in terms of language, culture, and acted vs.\ naturalistic speech is not
publicly documented, constituting a known limitation for cross-cultural
application.

Text-independent SER evaluation has been identified as substantially harder than
speaker-independent evaluation~\cite{atmaja2022text}; the gap is typically
10--20 percentage points in WA. Crucially, EMO-DB uses ten fixed sentences
spoken by all actors, making genuine text-independent evaluation structurally
impossible: any test sentence has appeared during training.

\subsection{LLM-Based Affect Analysis}

Recent work has explored LLMs as emotion annotators, with findings suggesting
that LLMs match or exceed crowdsourced human annotation on categorical emotion
tasks~\cite{amin2023will}. Gemini's multimodal architecture enables joint
processing of audio and text, allowing the model to integrate prosodic,
semantic, and contextual cues simultaneously. Unlike forced-choice paradigms
used in most SER corpora, open-ended LLM annotation avoids \textit{demand characteristics}---the tendency of annotators to select one of the presented options even when none fits well, leading to artificially inflated agreement with predefined categories.

\subsection{Computational Political Communication}

The TRUST pipeline~\cite{dietrich2026trust,dietrich2026roles} operationalises
Pathos as the societal impact of emotional language: from $+2$ (unifying across
party lines) through $0$ (neutral or group-internal) to $-2$ (actively
divisive). This definition differs substantially from the valence-arousal
circumplex used in affective computing, raising the question of which
computational approach better approximates TRUST-Pathos.

\section{Methods}
\label{sec:methods}

This section describes the data, models, and analysis pipeline used in this
study. All components are part of the TRUST Multimodal Pipeline (v1.0), an
open research system developed at Democracy Intelligence gGmbH.

\subsection{Speech Data}

We analyse a Bundestag plenary speech delivered by Felix Banaszak
(B\"{u}ndnis\,90/Die\,Gr\"{u}nen, co-chair), recorded on 5 March 2026 during
the 62nd session of the 21st German Bundestag (agenda item ZP\,3/4, energy
policy). Banaszak was selected as a representative case of oppositional
parliamentary rhetoric in the current German legislative period: as co-chair of
a party transitioning from government to opposition, his speech combines high
emotional intensity with complex rhetorical strategies (sarcasm, irony, appeal),
providing a demanding test case for emotion recognition systems. The study is
intentionally limited to a single speaker to enable controlled comparison of
modalities; multi-speaker generalisation is addressed in ongoing work. The speech was retrieved as a Full HD video stream
(1920$\times$1080, 8000\,kbps, H.264) from the official Bundestag media
library~\cite{bundestag2026} and converted to a mono 16\,kHz WAV file using
FFmpeg, excluding the first 12 seconds of procedural opening. Total duration:
232\,s. The speech was segmented into 51 utterances using
WhisperX~\cite{bain2023whisperx} with pyannote speaker
diarization~\cite{plaquet2023powerset}. Segment boundaries were determined by
pause-based and syntactic criteria, yielding utterances of typically 3--15
seconds.

\subsection{TRUST-Pathos Scoring}

Each segment was submitted to the TRUST pipeline in API mode. TRUST employs
three advocate LLMs---\textit{gemini-2.5-flash} (critical),
\textit{gpt-5.2} (balanced), and \textit{claude-sonnet-4-6}
(benevolent)---whose Pathos scores are aggregated by a supervisor LLM using
median consensus. Pathos scores are integers on a five-point scale:
$\{-2, -1, 0, +1, +2\}$, where $-2$ denotes actively divisive and $+2$
denotes societally unifying emotional language. Of the 51 segments, 10 were
excluded by the TRUST relevance filter (procedural utterances, greetings, and
closing statement), leaving 41 segments with valid Pathos scores.

\subsection{Acoustic Emotion Analysis: emotion2vec}

Acoustic emotion features were extracted using \textit{emotion2vec\_plus\_large}
(FunASR implementation~\cite{gao2023funasr}) at utterance granularity. The
model outputs class probabilities over eight categories: \textit{angry,
disgusted, fearful, happy, neutral, other, sad, surprised}.

\paragraph{Post-hoc Russell Circumplex Projection.}
emotion2vec does not natively output continuous Arousal or Valence values. We
derive these via a weighted sum of class probabilities, using weights adapted
from Russell~\cite{russell1980circumplex} and
Warriner et al.~\cite{warriner2013norms}:
\begin{equation}
  \text{Arousal} = \sum_{k} p_k \cdot w_k^{A}
  \label{eq:arousal}
\end{equation}
\begin{equation}
  \text{Valence} = \sum_{k} p_k \cdot w_k^{V}
  \label{eq:valence}
\end{equation}
where $p_k$ is the predicted probability for class $k$ and $w_k^{A}$,
$w_k^{V}$ are the Arousal and Valence weights listed in
Table~\ref{tab:weights}. This projection rests on three unverified assumptions:
(1) Russell weights transfer to German language; (2) they apply to naturalistic
political speech; (3) emotion2vec categories map onto Circumplex dimensions.
None of these assumptions has been empirically validated.

\begin{table}[ht]
\centering
\caption{Arousal and Valence weights for post-hoc Russell Circumplex projection
of emotion2vec class probabilities. Sources: Russell~\cite{russell1980circumplex};
Warriner et al.~\cite{warriner2013norms}.}
\label{tab:weights}
\begin{tabular}{lrr}
\toprule
\textbf{Class} & \textbf{$w^{A}$} & \textbf{$w^{V}$} \\
\midrule
angry     &  0.75 & $-$0.75 \\
disgusted &  0.60 & $-$0.80 \\
fearful   &  0.80 & $-$0.65 \\
happy     &  0.65 &    0.90 \\
neutral   &  0.00 &    0.00 \\
other     &  0.10 &    0.00 \\
sad       & $-$0.30 & $-$0.85 \\
surprised &  0.70 &    0.20 \\
\bottomrule
\end{tabular}

\smallskip
\noindent\textit{Abbreviations: $w^{A}$ = Arousal weight; $w^{V}$ = Valence weight.}
\end{table}

\subsection{LLM-Based Multimodal Analysis: Gemini}

We submitted the full speech audio together with the complete 51-segment
transcript (including segment IDs and timestamps) to \textit{Gemini 2.5 Flash}
(model: \texttt{gemini-2.5-flash}) via the Google GenAI API~(v1.74.0). The
system prompt instructed the model to evaluate each segment in terms of
(a)~primary and secondary emotion (open-ended, no forced choice),
(b)~Arousal on $[-1, +1]$, (c)~Valence on $[-1, +1]$,
(d)~rhetorical function (open-ended), and (e)~confidence on $[0, 1]$.
No predefined emotion categories were supplied---the model named emotions freely.
This open-ended paradigm avoids the demand characteristics inherent in
forced-choice annotation.

\subsection{Statistical Analysis}

Correlations between Arousal/Valence estimates from different modalities and
TRUST-Pathos scores were computed using the Spearman rank correlation
coefficient ($\rho$), appropriate for the ordinal TRUST-Pathos scale.
Statistical significance was assessed at $\alpha = 0.05$.

\section{Results}
\label{sec:results}

This section reports all empirical findings in two parts. Section~\ref{sec:emodb}
presents the EMO-DB quality evaluation, establishing Gemini's annotation
behaviour on acted German speech and identifying structural corpus limitations.
Section~\ref{sec:banaszak} presents the main comparative analysis on the
Banaszak plenary speech.

\subsection{EMO-DB Quality Evaluation}
\label{sec:emodb}

We evaluate Gemini as an open-ended annotator on all 535 EMO-DB utterances.
This serves two purposes: it characterises Gemini's emotion recognition on
acted German speech, and it reveals structural limitations of this benchmark
corpus. Appendix~A provides the complete Speaker$\times$Emotion matrix,
which also documents discrepancies between the published EMO-DB
documentation~\cite{burkhardt2005database} and the actual corpus files,
including differences in gender coding and sentence transcription found during
our manual listening evaluation.

\subsubsection{Corpus Structure}

EMO-DB comprises 535 utterances from ten speakers (six female: IDs 03, 08, 09,
10, 11, 12; four male: IDs 13, 14, 15, 16) across seven emotion categories
(Anger, Boredom, Disgust, Fear, Happiness, Neutral, Sadness) and ten fixed
German sentences. The distribution is imbalanced: Anger is the most frequent
class ($n = 127$, 23.7\,\%), Disgust the least frequent ($n = 46$, 8.6\,\%).
Speaker~08 produced no Disgust utterances, and speaker~09 produced only one
Fear utterance, creating systematic gaps in the Speaker$\times$Emotion matrix.

\subsubsection{Gemini Open-Ended Annotation}

Each WAV file was submitted individually to Gemini without predefined category
options. Gemini returned a primary emotion label, a secondary label
(or \texttt{null}), confidence ($[0,1]$), recording quality ($[0,1]$), and a
brief acoustic justification. Ground-truth (GT) matching used semantic
mapping: e.g., \textit{Sachlichkeit} (factuality) was mapped to Neutral,
\textit{Ver\"{a}rgerung} (annoyance) to Anger. Table~\ref{tab:emodb} reports
match rates per emotion category.

\begin{table}[ht]
\centering
\caption{Gemini open-ended annotation results on EMO-DB ($n = 535$). Match
rates reflect semantic matching between Gemini primary labels and EMO-DB GT.}
\label{tab:emodb}
\begin{tabular}{lrrr}
\toprule
\textbf{Emotion} & \textbf{$n$} & \textbf{Match (\%)} & \textbf{Avg.\ Conf.} \\
\midrule
Neutral   & 79  & 65.8 & 0.83 \\
Sadness   & 62  & 35.5 & 0.80 \\
Happiness & 71  & 29.6 & 0.83 \\
Anger     & 127 & 29.1 & 0.86 \\
Fear      & 69  & 27.5 & 0.77 \\
Boredom   & 81  & 12.3 & 0.81 \\
Disgust   & 46  &  0.0 & 0.81 \\
\midrule
\textbf{Total} & \textbf{535} & \textbf{30.1} & \textbf{0.82} \\
\bottomrule
\end{tabular}

\smallskip
\noindent\textit{Abbreviations: $n$ = number of utterances; Avg.\ Conf. =
mean Gemini confidence score; GT = ground truth.}
\end{table}

\subsubsection{Key Findings}

Three findings from Table~\ref{tab:emodb} merit discussion.

\textbf{Disgust: 0.0\,\% match.} Gemini consistently fails to identify Disgust
as a distinct category, labelling these utterances as annoyance, contempt, or
resignation. This suggests that Disgust is not acoustically distinguishable
without visual context (facial expression).

\textbf{Boredom: 12.3\,\% match.} Boredom is systematically misidentified as
Neutral or factual speech. Notably, Gemini's confidence is high (0.81) when
mislabelling Boredom utterances---confident but wrong. Manual listening by the
author confirmed that Boredom is readily identifiable to human listeners,
suggesting a representation gap for low-arousal German speech in Gemini's
training data.

\textbf{High confidence, low match.} The overall pattern---mean confidence 0.82
yet only 30.1\,\% semantic match---indicates that Gemini's confidence is a
poor predictor of correctness. This is consistent with a conceptual mismatch
between EMO-DB's forced-choice taxonomy and Gemini's open-ended vocabulary.

\subsubsection{Text Independence}

EMO-DB uses ten fixed sentences spoken by all actors, making genuine
text-independent evaluation structurally impossible. Our manual evaluation
further revealed that emotion-specific prosodic patterns are confounded with
sentence identity: Sadness is consistently produced with longer pauses across
all speakers, while Anger is produced staccato. A model can exploit these
text-specific rhythmic cues without learning emotion universals, inflating
text-dependent accuracy estimates.

\subsection{Banaszak Speech Analysis}
\label{sec:banaszak}

This section presents the comparative analysis on the 51-segment plenary
speech~\cite{bundestag2026}. We report descriptive statistics, Spearman
correlations, temporal dynamics, and Gemini's rhetorical classification.

\subsubsection{Descriptive Statistics}

Table~\ref{tab:descriptive} summarises Arousal and Valence estimates from both
modalities. Gemini assigns substantially higher Arousal (mean $= 0.59$) than
emotion2vec (mean $= 0.36$) and strongly negative Valence (mean $= -0.56$),
whereas emotion2vec Valence is near zero (mean $= 0.04$). TRUST-Pathos scores
cluster at $-1$ ($n = 18$) and $0$ ($n = 22$), with one segment at $-2$ and
one at $+1$.

\begin{table}[ht]
\centering
\caption{Descriptive statistics for Arousal, Valence, and TRUST-Pathos across
51 segments of the Banaszak speech~\cite{bundestag2026}.}
\label{tab:descriptive}
\begin{tabular}{lrrrr}
\toprule
\textbf{Measure} & \textbf{Mean} & \textbf{SD} & \textbf{Min} & \textbf{Max} \\
\midrule
Gemini Arousal      &  0.59 & 0.28 &  0.00 &  1.00 \\
Gemini Valence      & $-$0.56 & 0.44 & $-$1.00 &  0.60 \\
emotion2vec Arousal &  0.36 & 0.21 &  0.04 &  0.75 \\
emotion2vec Valence &  0.04 & 0.32 & $-$0.74 &  0.78 \\
TRUST-Pathos        & $-$0.37 & 0.56 & $-$2.00 &  1.00 \\
\bottomrule
\end{tabular}

\smallskip
\noindent\textit{Abbreviations: SD = standard deviation; TRUST = Transparent
Rhetorical Understanding and Scoring Tool.}
\end{table}

\subsubsection{Correlation Analysis}

Table~\ref{tab:correlations} reports Spearman rank correlations between all
modality pairs. The central finding is a strong and significant correlation
between Gemini Valence and TRUST-Pathos ($\rho = +0.664$, $p < 0.001$), and
a moderate negative correlation between Gemini Arousal and TRUST-Pathos
($\rho = -0.535$, $p < 0.001$). By contrast, emotion2vec Valence shows no
significant association with TRUST-Pathos ($\rho = +0.097$, $p = 0.499$), and
emotion2vec Arousal correlates weakly and non-significantly ($\rho = -0.155$,
$p = 0.278$). Cross-modal agreement is also low: emotion2vec and Gemini
Arousal correlate at $\rho = +0.239$ ($p = 0.091$), and Valence at
$\rho = +0.200$ ($p = 0.159$), confirming that the two approaches capture
substantially different dimensions. Segment-level scores are listed in
Appendix~B.

\begin{table}[ht]
\centering
\caption{Spearman rank correlations between emotion modalities and TRUST-Pathos
for 51 segments of the Banaszak speech~\cite{bundestag2026}.}
\label{tab:correlations}
\begin{tabular}{lrr}
\toprule
\textbf{Comparison} & \textbf{$\rho$} & \textbf{$p$} \\
\midrule
\textbf{Gemini Valence $\leftrightarrow$ TRUST-Pathos}  & \textbf{0.664} & \textbf{$<$0.001} \\
\textbf{Gemini Arousal $\leftrightarrow$ TRUST-Pathos}  & \textbf{$-$0.535} & \textbf{$<$0.001} \\
e2v Valence $\leftrightarrow$ TRUST-Pathos     &  0.097 &  0.499 \\
e2v Arousal $\leftrightarrow$ TRUST-Pathos     & $-$0.155 &  0.278 \\
e2v Arousal $\leftrightarrow$ Gemini Arousal   &  0.239 &  0.091 \\
e2v Valence $\leftrightarrow$ Gemini Valence   &  0.200 &  0.159 \\
\bottomrule
\end{tabular}

\smallskip
\noindent\textit{Abbreviations: $\rho$ = Spearman rank correlation coefficient;
e2v = emotion2vec; TRUST = Transparent Rhetorical Understanding and Scoring
Tool. Bold rows indicate $p < 0.001$; remaining rows $p \geq 0.05$ (not significant).}
\end{table}

\subsubsection{Temporal Dynamics}

\begin{figure}[ht]
\centering
\begin{tikzpicture}
\begin{axis}[
  width=\linewidth, height=5.8cm,
  xlabel={Segment index},
  ylabel={Score},
  ymin=-1.2, ymax=1.2,
  xmin=0, xmax=52,
  legend style={at={(0.5,-0.28)}, anchor=north, legend columns=3, font=\small},
  grid=major, grid style={dotted,gray!40},
  tick label style={font=\small},
  label style={font=\small},
]
\addplot[color=teal, thick, smooth] coordinates {
(0,0.1)(1,0.1)(2,0.0)(3,0.2)(4,0.1)(5,0.0)(6,-0.7)(7,-0.3)(8,-0.7)(9,-0.2)
(10,-0.3)(11,-0.4)(12,-0.7)(13,-0.8)(14,-0.3)(15,-0.4)(16,-0.8)(17,-0.6)
(18,-0.9)(19,-0.2)(20,-0.9)(21,-0.3)(22,-0.9)(23,-0.5)(24,-0.9)(25,-0.7)
(26,-0.2)(27,-0.1)(28,-0.2)(29,-0.8)(30,-0.9)(31,-0.7)(32,-0.9)(33,-0.3)
(34,-0.6)(35,-0.3)(36,-0.4)(37,-0.3)(38,-0.6)(39,-0.4)(40,-0.5)(41,-0.7)
(42,0.4)(43,-0.2)(44,-0.8)(45,-0.3)(46,-0.8)(47,-0.8)(48,0.1)(49,0.6)(50,0.0)
};
\addlegendentry{Gemini Valence}
\addplot[color=blue, dashed, thin] coordinates {
(0,0.12)(1,0.1)(2,0.12)(3,0.10)(4,0.16)(5,0.27)(6,0.18)(7,0.13)(8,0.09)
(9,0.31)(10,0.2)(11,0.23)(12,0.64)(13,0.45)(14,0.20)(15,0.23)(16,0.55)
(17,0.49)(18,0.65)(19,0.60)(20,0.69)(21,0.4)(22,0.75)(23,0.44)(24,0.41)
(25,0.33)(26,0.2)(27,0.04)(28,0.10)(29,0.44)(30,0.48)(31,0.42)(32,0.62)
(33,0.37)(34,0.23)(35,0.35)(36,0.50)(37,0.58)(38,0.54)(39,0.74)(40,0.46)
(41,0.13)(42,0.45)(43,0.53)(44,0.56)(45,0.45)(46,0.13)(47,0.05)(48,0.27)
(49,0.58)(50,0.3)
};
\addlegendentry{e2v Arousal}
\addplot[only marks, mark=*, mark size=3pt, color=violet] coordinates {
(3,0)(4,0)(5,0)(6,-1)(7,0)(8,-1)(9,0)(10,0)(11,0)(12,0)(13,-1)(14,0)
(15,0)(16,-1)(17,0)(18,-1)(19,0)(20,-1)(21,0)(22,-1)(23,0)(24,-1)(25,-1)
(26,0)(27,0)(28,0)(29,-1)(30,-2)(31,-1)(32,-1)(33,0)(34,-1)(35,0)(36,0)
(37,0)(38,-1)(39,0)(40,0)(41,-1)(42,1)(43,0)(44,-1)(45,0)(46,-1)(47,-1)(48,0)
};
\addlegendentry{TRUST Pathos}
\end{axis}
\end{tikzpicture}
\caption{Temporal profiles of Gemini Valence, emotion2vec (e2v) Arousal, and
TRUST-Pathos across 51 segments of the Banaszak speech~\cite{bundestag2026}.
Gemini Valence tracks the rhetorical arc; e2v Arousal reflects acoustic energy
independently of TRUST-Pathos. e2v Valence and Gemini Arousal are omitted for
clarity; both show no significant correlation with TRUST-Pathos
(Table~\ref{tab:correlations}).}
\label{fig:timeseries}

\smallskip
\noindent\textit{Abbreviations: e2v = emotion2vec; TRUST = Transparent
Rhetorical Understanding and Scoring Tool.}
\end{figure}
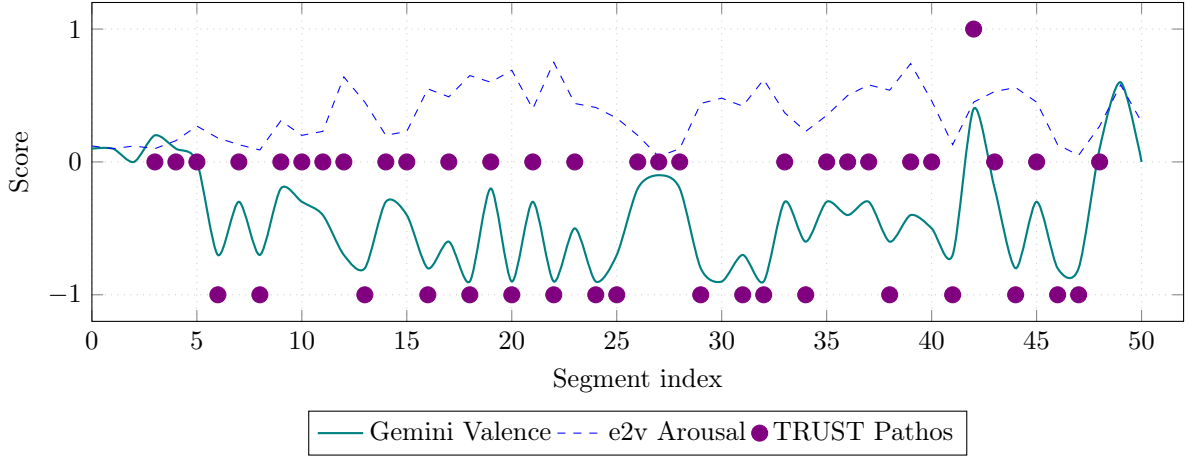

Figure~\ref{fig:timeseries} illustrates the temporal evolution of Gemini
Valence, emotion2vec Arousal, and TRUST-Pathos across the 51 segments.
Gemini Valence tracks the rhetorical arc of the speech closely, remaining
strongly negative throughout the main body (segments 6--47) and returning to
neutral only for the closing statement (segment~49). emotion2vec Arousal shows
high variability with no discernible correspondence to TRUST-Pathos. The single
positive TRUST-Pathos segment (s0042: \textit{``Es gibt einen Morgen---''},
``There is a tomorrow---'') coincides with the only positive Gemini Valence
value in the body of the speech, confirming that Gemini captures this
rhetorical turn correctly. Note that e2v Valence and Gemini Arousal are omitted
from the figure for clarity; their temporal profiles show no correspondence to
TRUST-Pathos (Table~\ref{tab:correlations}).

\subsubsection{Rhetorical Analysis}

Gemini's open-ended rhetorical classification reveals a distribution consistent
with oppositional parliamentary discourse: Criticism ($n = 16$, 31\,\%),
Sarcasm ($n = 14$, 27\,\%), None ($n = 9$, 18\,\%), Appeal ($n = 7$, 14\,\%),
Metaphor ($n = 2$, 4\,\%), Accusation ($n = 1$, 2\,\%), Rhetorical Question
($n = 1$, 2\,\%), and Indignation ($n = 1$, 2\,\%). This taxonomy---which
emerged without predefined categories---captures the rhetorical profile of an
opposition speech targeting the governing coalition, and corresponds to the
negative Pathos cluster observed in TRUST scoring.

\section{Discussion}
\label{sec:discussion}

Our results clarify why acoustic SER is insufficient as a Pathos proxy in
political communication analysis, and suggest a path forward.

\subsection{Two Different Constructs}

As shown in Table~\ref{tab:correlations}, the near-zero correlation between
emotion2vec Valence and TRUST-Pathos ($\rho = +0.097$, $p = 0.499$) is not a
failure of emotion2vec---it reflects the fact that the two measures capture
different constructs. emotion2vec captures \textit{acoustic Valence}: the
emotional colouring inferable from voice quality, fundamental frequency (F0)
contour, and spectral features. TRUST-Pathos captures
\textit{political-rhetorical Valence}: the societal impact of emotional
language, including irony, sarcasm, and rhetorical strategy accessible only
through semantic understanding.

The segment \textit{``Das ist wirklich peinlich''} (``That is truly
embarrassing'') illustrates this gap: emotion2vec assigns high positive Valence
($+0.74$) and classifies it as \textit{happy} based on acoustic energy. Gemini
correctly identifies the utterance as strong disapproval (Valence $-0.90$).
TRUST assigns Pathos $= 0$, reflecting that the statement is evaluatively
charged but directed at a specific target rather than broadly divisive.
More broadly, this points to a rhetorical strategy of \textit{decoupling}:
a speaker may deploy high prosodic activation to frame a statement
emotionally while delivering the propositional core in a factually neutral
register---a pattern that acoustic models cannot detect but LLMs can
identify through semantic understanding.

\subsection{The Post-Hoc Projection Problem}

Equations~(\ref{eq:arousal}) and~(\ref{eq:valence}) operationalise a
theoretically motivated but empirically unvalidated mapping. The Warriner
et al.~\cite{warriner2013norms} norms were derived from English word ratings,
not spoken German political discourse. The Russell
Circumplex~\cite{russell1980circumplex} is a model of subjective affect, not
of acoustic signal properties. Future work should empirically validate or
replace this projection using manual Arousal/Valence annotations of the target
corpus.

\subsection{EMO-DB Limitations}

Our evaluation reveals three concerns beyond those previously documented. First,
Disgust is acoustically unidentifiable without visual context. Second, Boredom
is systematically misclassified by Gemini despite being readily identifiable to
human listeners. Third, emotion-specific prosodic conventions create
text-specific cues that inflate text-dependent accuracy estimates.

\subsection{Limitations}

This study has several limitations. First, $n = 51$ segments from a single
speaker limits statistical power; we plan to extend the corpus to additional
speakers and speech types. Second, Gemini Arousal and Valence are
self-estimated scalar values derived from the same model that produces
rhetorical labels, creating potential internal consistency effects. Third,
TRUST-Pathos scores reflect a specific operationalisation of political emotion
developed and evaluated on German political statements~\cite{dietrich2026trust}.
Whether this operationalisation generalises to other political systems,
languages, or speech genres remains an open empirical question. Fourth, the
training data composition of emotion2vec is not publicly documented.

\section{Conclusion}
\label{sec:conclusion}

We have shown that LLM-based multimodal emotion analysis substantially
outperforms acoustic SER as a proxy for TRUST-Pathos in naturalistic political
speech. The key mechanism is semantic-pragmatic understanding: Gemini integrates
lexical content, rhetorical structure, and political context, whereas emotion2vec
responds primarily to acoustic signal properties. Both modalities capture real
information---but about different dimensions of political communication.

These results have practical implications for the TRUST pipeline and for
computational political communication analysis more broadly. Rather than
replacing acoustic analysis, LLM-based emotion scoring should be used in
conjunction with acoustic features, creating a complementary multimodal
representation of political affect. Future work will extend this analysis to a
multi-speaker corpus, evaluate fine-tuned SER models (wav2vec\,2.0 trained on
quality-filtered EMO-DB and PAVOQUE), investigate whether such a model better approximates TRUST-Pathos than the
base acoustic model, and compare multiple LLM providers (e.g., Gemini, GPT,
Claude) as emotion annotators to assess whether the observed correlation with
TRUST-Pathos is model-specific or reflects a more general property of
semantically informed emotion analysis. A natural further
extension is the incorporation of video analysis---combining facial expression
recognition (Action Units via OpenFace), gaze estimation (L2CS-Net), and body
posture tracking (MediaPipe)---to capture the full multimodal dimension of
political communication.

\section*{Acknowledgements}

The author thanks Demian Frister (Democracy Intelligence gGmbH) for his
critical review and valuable comments on an earlier version of this manuscript.

\clearpage

\bibliographystyle{unsrt}
\bibliography{paper8}

\clearpage

\appendix

\section{Appendix A: EMO-DB Speaker$\times$Emotion Matrix}
\label{app:emodb_matrix}

Table~\ref{tab:emodb_matrix} shows the complete distribution of utterances
across speakers and emotion categories in EMO-DB. The matrix reveals systematic
gaps---most notably, speaker~08 produced no Disgust utterances---that complicate
speaker-independent cross-validation. A manual review of 140 of the 535 files revealed discrepancies between
the published EMO-DB documentation~\cite{burkhardt2005database} and the actual
corpus files, including differences in gender coding and sentence transcription.
A complete verification against the original recording notes is recommended for
future users of the corpus.

\begin{table}[ht]
\centering
\caption{EMO-DB utterance counts per speaker and emotion category.
Speaker and emotion distribution as derived from filename metadata.}
\label{tab:emodb_matrix}
\begin{tabular}{lcrrrrrrrr}
\toprule
\textbf{Speaker} & \textbf{G} & \textbf{Ang} & \textbf{Bor} &
\textbf{Dis} & \textbf{Fea} & \textbf{Hap} & \textbf{Neu} &
\textbf{Sad} & \textbf{Total} \\
\midrule
03 & F & 14 &  5 &  1 &  4 &  7 & 11 &  7 & 49 \\
08 & F & 12 & 10 &  0 &  6 & 11 & 10 &  9 & 58 \\
09 & F & 13 &  4 &  8 &  1 &  4 &  9 &  4 & 43 \\
10 & F & 10 &  8 &  1 &  8 &  4 &  4 &  3 & 38 \\
11 & F & 11 &  8 &  2 & 10 &  8 &  9 &  7 & 55 \\
12 & F & 12 &  5 &  2 &  6 &  2 &  4 &  4 & 35 \\
13 & M & 12 & 10 &  8 &  7 & 10 &  9 &  5 & 61 \\
14 & M & 16 &  8 &  8 & 12 &  8 &  7 & 10 & 69 \\
15 & M & 13 &  9 &  5 &  8 &  6 & 11 &  4 & 56 \\
16 & M & 14 & 14 & 11 &  7 & 11 &  5 &  9 & 71 \\
\midrule
\textbf{Total} & & 127 & 81 & 46 & 69 & 71 & 79 & 62 & 535 \\
\bottomrule
\end{tabular}

\smallskip
\noindent\textit{Abbreviations: G = gender (F = female, M = male);
Ang = Anger; Bor = Boredom; Dis = Disgust; Fea = Fear; Hap = Happiness;
Neu = Neutral; Sad = Sadness.}
\end{table}

\clearpage

\section{Appendix B: Banaszak Speech -- Segment-Level Scores}
\label{app:segments}

Table~\ref{tab:segments} lists the 41 segments retained for analysis after
applying the TRUST relevance filter. Ten segments were excluded as they
constitute non-evaluable utterances: procedural address to the chair, closing
statement, and moderator announcements. The complete speech is publicly
available~\cite{bundestag2026}.

\begin{landscape}
\begin{table}[p]
\centering
\tiny
\caption{Segment-level scores for the Banaszak speech~\cite{bundestag2026},
41 analysed segments. Pathos scores are integers on $\{-2,-1,0,+1,+2\}$.
\textit{Abbreviations: e2v-A = emotion2vec Arousal; e2v-V = emotion2vec Valence;
Gem-A = Gemini Arousal; Gem-V = Gemini Valence;
Pathos = TRUST-Pathos score; Gem-Emotion = Gemini primary emotion (German);
Gem-Rhetoric = Gemini rhetorical function (English).}}
\label{tab:segments}
\begin{tabular}{llrrrrrp{2.2cm}p{1.9cm}}
\toprule
\textbf{ID} & \textbf{Text (abridged)} &
\textbf{e2v-A} & \textbf{e2v-V} &
\textbf{Gem-A} & \textbf{Gem-V} &
\textbf{Pathos} & \textbf{Gem-Emotion} & \textbf{Gem-Rhetoric} \\
\midrule
s0003 & Die Klima- und Umweltbewegung\ldots  & 0.10 & $-$0.07 & 0.50 &  0.20 &  0 & Entschlossenheit & Appell \\
s0004 & \"{U}ber 200\,000 Unterschriften     & 0.16 &  0.21 & 0.40 &  0.10 &  0 & Sachlichkeit     & keines \\
s0005 & In dieser Debatte\ldots              & 0.27 &  0.36 & 0.50 &  0.00 &  0 & Begeisterung     & Appell \\
s0006 & 200\,000 Unterschriften\ldots        & 0.18 &  0.22 & 0.60 & $-$0.70 & $-$1 & Verachtung   & Kritik \\
s0007 & Lassen Sie mich\ldots               & 0.13 &  0.12 & 0.40 & $-$0.20 &  0 & Entschlossenheit & keines \\
s0008 & 200\,000 Menschen\ldots             & 0.09 &  0.00 & 0.60 & $-$0.70 & $-$1 & Spott        & Sarkasmus \\
s0009 & Die drei von der Tankstelle\ldots   & 0.31 &  0.33 & 0.50 & $-$0.20 &  0 & Sarkasmus        & Sarkasmus \\
s0013 & Aber dass Jens Spahn\ldots          & 0.46 &  0.06 & 0.70 & $-$0.80 & $-$1 & Emp\"{o}rung & Emp\"{o}rung \\
s0014 & Aber das System\ldots               & 0.20 &  0.27 & 0.50 & $-$0.30 &  0 & Kritik           & Kritik \\
s0015 & Das Hab\"{a}cksche\ldots            & 0.23 &  0.27 & 0.50 & $-$0.40 &  0 & Sarkasmus        & Sarkasmus \\
s0016 & Und Mirsch erz\"{a}hlt\ldots        & 0.55 & $-$0.04 & 0.70 & $-$0.80 & $-$1 & Kritik      & Sarkasmus \\
s0017 & Nur weil man\ldots                  & 0.49 & $-$0.36 & 0.60 & $-$0.60 &  0 & Frustration      & Kritik \\
s0018 & Heute w\"{a}re die Chance\ldots     & 0.65 &  0.28 & 0.80 & $-$0.90 & $-$1 & Anklage      & Kritik \\
s0019 & Heute w\"{a}re die Chance\ldots     & 0.60 &  0.72 & 0.50 & $-$0.20 &  0 & Vorwurf          & Kritik \\
s0020 & Solaranlage\ldots                   & 0.69 & $-$0.29 & 0.80 & $-$0.90 & $-$1 & Emp\"{o}rung & Kritik \\
s0022 & Jetzt habt ihr\ldots                & 0.75 & $-$0.74 & 0.80 & $-$0.90 & $-$1 & Sarkasmus    & Sarkasmus \\
s0023 & Also gibt es keine\ldots            & 0.44 & $-$0.38 & 0.60 & $-$0.50 &  0 & Kritik           & Kritik \\
s0024 & Heute w\"{a}re die Chance\ldots     & 0.41 & $-$0.32 & 0.80 & $-$0.90 & $-$1 & Anklage      & Appell \\
s0025 & Sie lassen diese Chance\ldots       & 0.33 & $-$0.03 & 0.70 & $-$0.70 & $-$1 & Vorwurf      & Kritik \\
s0027 & Katharina Reich hat\ldots           & 0.04 &  0.02 & 0.30 & $-$0.10 &  0 & Sachlichkeit     & keines \\
s0028 & Ja, das w\"{u}rde ich\ldots         & 0.10 &  0.05 & 0.40 & $-$0.20 &  0 & Sarkasmus        & Sarkasmus \\
s0029 & Stattdessen\ldots Traumabew\ldots   & 0.44 &  0.30 & 0.70 & $-$0.80 & $-$1 & Verachtung   & Kritik \\
s0030 & Alles, was nach Habeck\ldots        & 0.48 & $-$0.21 & 0.80 & $-$0.90 & $-$2 & Emp\"{o}rung & Sarkasmus \\
s0031 & Das ist doch nicht Freiheit         & 0.42 &  0.01 & 0.70 & $-$0.70 & $-$1 & Emp\"{o}rung & Rhet.\ Question \\
s0032 & Sie bringen Energiearmut\ldots      & 0.62 &  0.45 & 0.80 & $-$0.90 & $-$1 & Emp\"{o}rung & Anklage \\
s0033 & Seit 2022 wissen wir\ldots          & 0.37 &  0.02 & 0.50 & $-$0.30 &  0 & Sachlichkeit     & Appell \\
s0034 & Sie haben nichts gelernt            & 0.23 & $-$0.07 & 0.60 & $-$0.60 & $-$1 & Vorwurf      & Kritik \\
s0035 & Es gibt kein Verbot\ldots           & 0.35 & $-$0.03 & 0.50 & $-$0.30 &  0 & Sarkasmus        & Sarkasmus \\
s0036 & Mit Verboten kenne ich\ldots        & 0.50 &  0.01 & 0.50 & $-$0.40 &  0 & Sarkasmus        & Sarkasmus \\
s0037 & Es gibt doch kein Verbot            & 0.58 & $-$0.02 & 0.50 & $-$0.30 &  0 & Kritik           & Kritik \\
s0038 & Es gibt kein Gesetz\ldots           & 0.54 &  0.14 & 0.70 & $-$0.60 & $-$1 & Kritik       & Kritik \\
s0039 & Und da ist die Wand\ldots           & 0.74 & $-$0.66 & 0.60 & $-$0.40 &  0 & Metapher         & Metapher \\
s0040 & Jetzt stehen Sie vor\ldots          & 0.46 & $-$0.25 & 0.60 & $-$0.50 &  0 & Kritik           & Kritik \\
s0041 & Das ganze Land steht\ldots          & 0.13 & $-$0.04 & 0.70 & $-$0.70 & $-$1 & Metapher     & Metapher \\
s0042 & Es gibt einen Morgen\ldots          & 0.45 & $-$0.30 & 0.50 &  0.40 &   1 & Zuversicht        & Appell \\
s0043 & Robin Mesarosch\ldots               & 0.53 &  0.35 & 0.50 & $-$0.20 &  0 & Sarkasmus        & Sarkasmus \\
s0044 & Als h\"{a}tten Sie\ldots            & 0.56 & $-$0.15 & 0.70 & $-$0.80 & $-$1 & Sarkasmus    & Sarkasmus \\
s0045 & Das ist Ihr Fraktions\ldots         & 0.45 & $-$0.23 & 0.50 & $-$0.30 &  0 & Kritik           & Kritik \\
s0046 & Ich sage Ihnen\ldots                & 0.13 &  0.13 & 0.70 & $-$0.80 & $-$1 & Emp\"{o}rung & Appell \\
s0047 & Zeigen Sie Gr\"{o}\ss e\ldots       & 0.05 &  0.02 & 0.90 & $-$0.80 & $-$1 & Appell       & Appell \\
s0048 & auf unsere Sicherheit\ldots         & 0.27 &  0.33 & 0.40 &  0.10 &  0 & Appell            & Appell \\
\bottomrule
\end{tabular}

\end{table}
\end{landscape}

\end{document}